\documentclass[11pt,letterpaper]{article}
\usepackage{naaclhlt2016}
\usepackage{times}
\usepackage{latexsym}

\usepackage{mystuff-nlp}
\usepackage{url}
\usepackage{color}
\usepackage{float}
\makeatletter
\newcommand{\@BIBLABEL}{\@emptybiblabel}
\newcommand{\@emptybiblabel}[1]{}
\makeatother
\usepackage[draft,hidelinks]{hyperref}
\usepackage{amsfonts,amssymb,amsmath}
\usepackage{algorithm}
\usepackage{appendix}
\usepackage{algpseudocode}
\usepackage{graphicx,subfigure}
\usepackage{booktabs}
\usepackage{multirow}

\usepackage{enumitem}
\naaclfinalcopy 
\def\naaclpaperid{278} 

\newcommand{\mnote}[1]{\marginpar{%
\vskip-\baselineskip
\raggedright\footnotesize
\itshape\hrule\smallskip\tiny{\color{red}{#1}}\par\smallskip\hrule}}

\newcommand{\mb}[1]{\mathbf{#1}}

\newcommand{\sysname}{\textsc{Fema}}

\newcommand{\example}[1]{`#1'}

\newcommand{\postag}[1]{\texttt{#1}}

\title{Part-of-Speech Tagging for Historical English}

\author{Yi Yang \and Jacob Eisenstein\\
	    School of Interactive Computing\\
	    Georgia Institute of Technology\\
            Atlanta, GA 30308\\
	    {\tt \{yiyang+jacobe\}@gatech.edu}}

\date{}

\begin{document}
\maketitle
\begin{abstract}
As more historical texts are digitized, there is interest in applying natural language processing tools to these archives. However, the performance of these tools is often unsatisfactory, due to language change and genre differences. Spelling normalization heuristics are the dominant solution for dealing with historical texts, but this approach fails to account for changes in usage and vocabulary. In this empirical paper, we assess the capability of domain adaptation techniques to cope with historical texts, focusing on the classic benchmark task of part-of-speech tagging. We evaluate several domain adaptation methods on the task of tagging Early Modern English and Modern British English texts in the Penn Corpora of Historical English. We demonstrate that the Feature Embedding method for unsupervised domain adaptation outperforms word embeddings and Brown clusters, showing the importance of embedding the entire feature space, rather than just individual words. Feature Embeddings also give better performance than spelling normalization, but the combination of the two methods is better still, yielding a 5\% raw improvement in tagging accuracy on Early Modern English texts.


\end{abstract}

\section{Introduction}
\label{sec:intro}

There is growing interest in applying natural language processing (NLP) techniques to historical texts~\cite{piotrowski2012natural}, with applications in information retrieval~\cite{dougherty2010google,jurish2011finite}, linguistics~\cite{baron2009word,rayson2007tagging}, and the digital humanities~\cite{hendrickx2011automatic,muralidharan2013supporting,pettersson2011automatic}. However, these texts differ from contemporary training corpora in a number of linguistic respects, including the lexicon~\cite{giusti2007automatic}, morphology~\cite{borin2008something}, and syntax~\cite{eumeridou2004analysis}. This imposes significant challenges for modern NLP tools: for example, the accuracy of the CLAWS part-of-speech Tagger~\cite{garside1997hybrid} drops from 97\% on the British National Corpus to 82\% on Early Modern English texts~\cite{rayson2007tagging}. There are two main approaches that could improve the accuracy of NLP systems on historical texts: normalization and domain adaptation.

\begin{figure}[t]
  \centering
   \begin{footnotesize}
   \framebox {
   \parbox[l]{.9 \columnwidth}{
   {\bf Original}: and drewe vnto hym all ryottours \& wylde dysposed persones \\
   {\bf Normalization}: and drew unto him all ryottours \& wild disposed persons
} }
   \end{footnotesize}
  \caption{An example sentence from Early Modern English and its VARD normalization.}\label{fig:example}
\end{figure}

\paragraph{Normalization} Spelling normalization (also called canonicalization) involves mapping historical spellings to their canonical forms in modern languages, thus bridging the gap between contemporary training corpora and target historical texts. Figure~\ref{fig:example} shows one historical sentence and its normalization by VARD~\cite{baron2008vard2}. 
\newcite{rayson2007tagging} report an increase of about 3\% accuracy on adaptation of POS tagging from Modern English texts to Early Modern English texts if the target texts were automatically normalized by the VARD system. However, normalization is not always a well-defined problem~\cite{eisenstein2013bad}, and it does not address the full range of linguistic changes over time, such as unknown words, morphological differences, and changes in the meanings of words~\cite{kulkarni2015statistically}. In the example above, the word \example{ryottours} is not successfully normalized to \example{rioters}; the syntax is comprehensible to contemporary English speakers, but usages such as \example{wild disposed} and \example{drew unto} are sufficiently unusual as to pose problems for NLP systems trained on contemporary texts.

\paragraph{Domain adaptation} A more generic machine learning approach is to apply unsupervised domain adaptation techniques, which transform the representations of the training and target texts to be more similar, typically using feature co-occurrence statistics~\cite{blitzer2006domain,ben2010theory}. It is natural to think of historical texts as a distinct domain from contemporary training corpora, and Yang and Eisenstein (2014, 2015)\nocite{yang2014fast,yang2015unsupervised} show that the accuracy of historical Portuguese POS tagging can be significantly improved by domain adaption. However, we are unaware of prior work that empirically evaluates the efficacy of this approach on Early Modern English texts. 
Furthermore, historical texts are often associated with multiple metadata attributes (e.g., author, genre, and epoch), each of which may influence the text's linguistic properties. \emph{Multi-domain adaptation}~\cite{mansour2009domain} and \emph{multi-attribute domain adaptation}~\cite{joshi2013what,yang2015unsupervised} can potentially exploit these metadata attributes to obtain further improvements. 

This paper presents the first comprehensive empirical comparison of effectiveness of these approaches for part-of-speech tagging on historical texts. We focus on the two historical treebanks of the Penn Corpora of Historical English --- the Penn Parsed Corpus of Modern British English~\cite[PPCMBE]{kroch2010penn} and the Penn-Helsinki Parsed Corpus of Early Modern English~\cite[PPCEME]{kroch2004penn}. These datasets enable a range of analyses, which isolate the key issues in dealing with historical corpora:
\begin{itemize}[itemsep=0pt]
\item In one set of analyses, we focus on the PPCMBE and the PPCEME corpora, training on more recent texts and testing on earlier texts. This isolates the impact of language change on tagging performance.
\item In another set of analyses, we train on the Penn Treebank~\cite[PTB]{marcus1993building}, and test on the historical corpora, using the tag mappings from \newcite{moon2007part}. We apply the well-known Stanford CoreNLP tagger to this task~\cite{manning2014stanford}, thus replicating the most typical situation for users of existing language technology.
\item We show that \sysname, a domain adaptation algorithm that is specifically designed for sequence labeling problems~\cite{yang2015unsupervised}, achieves an increase of nearly 4\% in tagging accuracy when adapting from the PTB to the PPCEME.
\item We compare the impact of normalization with domain adaptation, and demonstrate that they are largely complementary.
\item Error analysis shows that the improvements obtained by domain adaptation are largely due to better handling of out-of-vocabulary (OOV) tokens. Many of the most frequent errors on in-vocabulary (IV) tokens are caused by mismatches in the tagsets or annotation guidelines, and may be difficult to address without labeled data in the target domain.
\end{itemize}

\section{Data}
\label{sec:data}

The Penn Corpora of Historical English consist of the Penn-Helsinki Parsed Corpus of Middle English, second edition~\cite[PPCME2]{kroch2010penn}, the Penn-Helsinki Parsed Corpus of Early Modern English~\cite[PPCEME]{kroch2004penn}, and the Penn Parsed Corpus of Modern British English~\cite[PPCMBE]{kroch2000penn}. The corpora are annotated with part-of-speech tags and syntactic parsing trees in an annotation style similar to that of the Penn Treebank. In this work, we focus on POS tagging the PPCMBE and the PPCEME.\footnote{Middle English is outside the scope of this paper, because it is sufficiently unintelligible to modern English speakers that texts such as Canterbury Tales are published in translation. 
In tagging Middle English texts, \newcite{moon2007part} apply bitext projection techniques from multilingual learning, rather than domain adaptation.}

\paragraph{The Penn Parsed Corpus of Modern British English}
The PPCMBE is a syntactically annotated corpus of text, containing 
roughly one million word tokens from documents written in the period 1700-1914. It is divided into three 70-year time periods according to the composition date of the works. \autoref{tab:ppcmbe-time} shows the statistics of the corpus by time period.\footnote{All the statistics in this section include punctuation, but exclude extra-linguistic material such as page numbers or token ID numbers.} In contrast to the PTB, the PPCMBE contains text from a variety of genres, such as Bible, Drama, Fiction, and Letters.

\begin{table} [t]
\centering
\small
\addtolength{\tabcolsep}{-2pt}
\begin{tabular}{lrr}
    \toprule
    Period & \# Sentence & \# Token \\ \midrule
    1840-1914 & 17,770 & 322,255 \\
    1770-1839 & 23,462 & 427,424 \\
    1700-1769 & 16,083 & 343,024 \\ \midrule
    Total     & 57,315 & 1,092,703 \\
    \bottomrule
\end{tabular}
\caption{Statistics of the Penn Parsed Corpus of Modern British English (PPCMBE), by time period.}
\label{tab:ppcmbe-time}
\end{table}

\paragraph{The Penn-Helsinki Parsed Corpus of Early Modern English}
The PPCEME is a collection of text samples from the Helsinki Corpus~\cite{rissanen1993early}, as well as two supplements mainly consisting of text material by the same authors and from the same editions as the material in the Helsinki Corpus. The corpus contains 
nearly two million words from texts in the period from 1500 until 1710, and it is divided into three 70-year time periods similar to the PPCMBE corpus.  The statistics of the corpus by time period is summarized in \autoref{tab:ppceme-time}. The PPCEME consists of text from the same eighteen genres as the PPCMBE.

\begin{table} [t]
\centering
\small
\addtolength{\tabcolsep}{-2pt}
\begin{tabular}{lrr}
    \toprule
    Period & \# Sentence & \# Token \\ \midrule
    1640-1710 & 29,181 & 614,315 \\
    1570-1639 & 39,799 & 706,587 \\
    1500-1569 & 31,416 & 640,255 \\ \midrule
    Total     & 100,396 & 1,961,157 \\
    \bottomrule
\end{tabular}
\caption{Statistics of the Penn Parsed Corpus of Early Modern English (PPCEME), by time period.}
\label{tab:ppceme-time}
\end{table}


\paragraph{Penn Treebank Release 3}
The Penn Treebank~\cite{marcus1993building} is the de facto standard syntactically annotated corpus for English, which is used to train software such as Stanford CoreNLP~\cite{manning2014stanford}. When using this dataset for supervised training, we follow~\newcite{toutanova2003feature} and use WSJ sections 0-18 for training, and sections 19-21 for tuning. When applying unsupervised domain adaptation, we use all WSJ sections, together with texts from the PPCMBE and the PPCEME.

\paragraph{Tagsets}
The Penn Corpora of Historical English (PCHE) use a tagset that differs from the Penn Treebank, mainly in the direction of greater specificity. Auxiliary verbs \example{do}, \example{have}, and \example{be} all have their own tags, as do words like \example{one} and \example{else}, due to their changing syntactic function over time. Overall, there are 83 tags in the PPCEME, and 81 in the PPCMBE, as compared with 45 in the PTB. Furthermore, the tags in the PCHE tagset are allowed to join constituent morphemes in compounds, yielding complex tags such as \postag{PRO+N} (e.g., \example{himself}) and \postag{ADJ+NS} (e.g., \example{gentlemen}). 

To measure the tagging accuracy of PTB-trained taggers on the historical texts, we follow \newcite{moon2007part}, who define a set of deterministic mappings from the PCHE tags to the PTB tagset. For simplicity, we first convert each complex tag to the simple form by only considering the first simple tag component (e.g., \postag{PRO+N} to \postag{PRO} and \postag{ADJ+NS} to \postag{ADJ}). This has little effect on the tagging performance, as the complex tags cover only slightly more than 1\% of the tokens in the PCHE treebanks.
Among the 83 tags, 74 mappings to the corresponding PTB tags are obtained from~\newcite{moon2007part}. We did our best to convert the other tags according to the tag description. The complete list of mappings is published in Appendix~\ref{app:mappings}.


\section{Unsupervised Domain Adaptation}
\label{sec:model}
In typical usage scenarios, the user wants to tag some historical text but has no labeled data in the target domain (e.g., Muralidharan and Hearst, 2013\nocite{muralidharan2013supporting}). This best fits the paradigm of unsupervised domain adaptation, when labeled data from the source domain (e.g., the PTB) is combined with unlabeled data from the target domain. Representational differences between source and target domains can be a major source of errors in domain adaptation~\cite{ben2010theory}, and so several representation learning approaches have been proposed. 

The most straightforward approach is to replace lexical features with \textbf{word representations}, such as Brown clusters~\cite{brown1992class,lin2012syntactic} or word embeddings~\cite{turian2010word}, such as word2vec~\cite{mikolov2013distributed}. Lexical features can then be replaced or augmented with the resulting word representations. This can assist in domain adaptation by linking out-of-vocabulary words to in-vocabulary words with similar distributional properties. 

Word representations are suitable for adapting lexical features, but a more general solution is to adapt the entire feature representation. One such method is \textbf{Structural Correspondence Learning}~\cite[\textbf{SCL}]{blitzer2006domain}. In SCL, we create artificial binary classification problems for thousands of cross-domain ``pivot'' features, and then use the weights from the resulting classifiers to project the instances into a new dense representation. We also consider a recently-published approach called \textbf{Feature Embedding} (\textbf{\sysname}), which achieves the state-of-the-art results on several POS tagging adaptation tasks~\cite{yang2015unsupervised}. The intuition of \sysname\ is similar to SCL and other prior work: it relies on co-occurrence statistics to link features across domains. Specifically, \sysname\ exploits the tendency of many NLP tasks to divide features into templates, and induces feature embeddings by using the features in each template to predict the active features in all other templates --- just as the skipgram model learns word embeddings to predict neighboring words. The resulting embeddings can be substituted for the ``one-hot'' representation of each feature template, resulting in a dense, low-dimensional representation of each instance.

A further advantage of \sysname\ is that it can perform multi-attribute domain adaptation, enabling it to exploit the many metadata attributes (e.g., year, genre, and author) that are often associated with historical texts. This is done by accounting for the specific impact of each domain attribute on the feature predictors, and then building a domain-neutral representation from the common substructure that is shared across all domain attributes. In the experiments that follow, we use genre and epoch as domain attributes. 

\section{Experiments}
We evaluate these unsupervised domain adaptation approaches on part-of-speech tagging for historical English (the PPCMBE and the PPCEME), in two settings: (1) temporal adaptation within each individual corpus, where we train POS taggers on the most modern data in the corpus and test on increasingly distant datasets; (2) adaptation of English POS tagging from modern news text to historical texts. The first setting focuses on temporal differences, and eliminates other factors that may impair tagging performance, such as different annotation schemes and text genres. The second setting is the standard and well-studied evaluation scenario for POS tagging, where we train on the Wall Street Journal (WSJ) text from the PTB and test on historical texts. In addition, we evaluate the effectiveness of the VARD normalization tool~\cite{baron2008vard2} for improving POS tagging performance on the PPCEME corpus.

\subsection{Experimental Settings}
The datasets used in the experiments are described in~\autoref{sec:data}. All the hyperparameters are tuned on development data in the source domain. In the case where there is no specific development dataset (adaptation within the historical corpora), we randomly sample 10\% sentences from the training datasets for hyperparameter tuning. 

\subsubsection{Baseline systems} 
We include two baseline systems for POS tagging: a classification-based support vector machine (SVM) tagger and a bidirectional maximum entropy Markov model (MEMM) tagger. Specifically, we use the $L_2$-regularized $L_2$-loss SVM implementation in the scikit-learn package~\cite{pedregosa2011scikit} and $L_2$-regularized bidirectional MEMM implementation provided by Stanford CoreNLP~\cite{toutanova2003feature,manning2014stanford}. 

Following~\newcite{yang2015unsupervised}, we apply the feature templates defined by~\newcite{ratnaparkhi1996maximum} to extract the basic features for all taggers.
There are three broad types of templates: five lexical feature templates, eight affix feature templates, and three orthographic feature templates. 

\begin{table*}
\centering
\addtolength{\tabcolsep}{-2pt}
\begin{tabular}{l l l l l l l l}
    \toprule
    \multirow{2}{*}{Task} & \multicolumn{2}{c}{baseline} & \multirow{2}{*}{SCL} & \multirow{2}{*}{Brown} & \multirow{2}{*}{word2vec}\hspace{0.0cm} & \multicolumn{2}{c}{\sysname}\\
\cmidrule{2-3} \cmidrule{7-8}
& SVM & \parbox{1.4cm}{MEMM\\(Stanford)} & & & & \parbox{1.6cm}{single\\embedding} & \parbox{3.4cm}{attribute embeddings\\(error reduction)} \\ \midrule
\multicolumn{7}{l}{\it Modern British English (training from 1840-1914)} \\
$\rightarrow$ 1770-1839 & 96.30 & 96.57 & 96.42 & 96.45 & 96.44 & 96.80 & \textbf{96.84} (15\%) \\
$\rightarrow$ 1700-1769 & 94.57 & 94.83 & 95.07 & 95.15 & 94.85 & 95.65 & \textbf{95.75} (22\%) \\ 
    \textsc{average}    & 95.43 & 95.70 & 95.74 & 95.80 & 95.64 & 96.23 & \textbf{96.30} (19\%) \\[6pt]
\multicolumn{7}{l}{\it Early Modern English (training from 1640-1710)}\\
$\rightarrow$ 1570-1639 & 93.62 & 93.98 & 94.23 & 94.36 & 94.18 & 95.01 & \textbf{95.20} (25\%) \\
$\rightarrow$ 1500-1569 & 87.59 & 87.47 & 89.39 & 89.73 & 89.30 & 91.40 & \textbf{91.63} (33\%) \\
    \textsc{average}    & 90.61 & 90.73 & 91.81 & 92.05 & 91.74 & 93.20 & \textbf{93.41} (30\%) \\
    \bottomrule
\end{tabular}
\caption{Accuracy results for temporal adaptation in the PPCMBE and the PPCEME of historical English. Percentage error reduction is shown for the best-performing method, \sysname-attribute.}
\label{tab:res-ppc}
\end{table*}

\subsubsection{Domain adaptation systems}
We consider the unsupervised domain adaptation methods described in~\autoref{sec:model}: structural correspondence learning (SCL), Brown clustering, word2vec,\footnote{https://code.google.com/p/word2vec/} and \sysname, which we train in both the single embedding mode (\sysname-single), where metadata attributes are ignored, and in multi-attribute mode (\sysname-attribute), where metadata attributes are used.
The domain adaptation models are trained on the union of the (unlabeled) source and target datasets. This ensures that there are no out-of-vocabulary items for the word or feature embeddings.

Following~\newcite{yang2015unsupervised}, we do not learn feature embeddings for the three orthographic feature templates: as each orthographic feature template represents only a binary value, it is unnecessary to replace it with a much longer numerical vector. The learned representations are then concatenated with the basic surface features to form the augmented representations. For computational reasons, the domain adaptation systems are all based on the SVM tagger, as pilot studies showed that Viterbi tagging offers minimal improvements.

\subsubsection{Parameter tuning}
We choose the SVM regularization parameter by sweeping the range $\{ 0.1, 0.3, 0.5, 0.8, 1.0 \}$. Following~\newcite{blitzer2006domain}, we consider pivot features that appear more than 50 times in all the domains for SCL. We empirically fix the number of singular vectors of the projection matrix $K$ to 25, and also employ feature normalization and rescaling, as these settings yield best performance in prior work. The number of Brown clusters is chosen from the range $\{50, 100, 200, 400\}$. For \sysname~and word2vec, we choose embedding sizes from the range $\{50, 100, 200, 300\}$ and fix the numbers of negative samples to 15. The window size for training word embeddings is set as 5. Finally, we adopt the same regularization penalty for all the attribute-specific embeddings of \sysname, which is selected from the range $\{0.01, 0.1, 1.0, 10.0\}$. All parameters were tuned on development data in the source domain. We train the Stanford MEMM tagger using the default configuration file.

\subsection{Temporal Adaptation}
In the temporal adaptation setting, we work within each corpus, training on the most recent section, and evaluating on the two earlier sections. For PPCMBE, the source domain is the period from 1840 to 1914; for PPCEME, the source domain is the period from 1640 to 1710. All earlier texts are treated as target domains. We transform the tags to the PTB tagset for evaluation, so that results can be compared with the next experiment, in which the PTB is used for supervision.

\paragraph{Settings} We randomly sample 10\% sentences from the training data as the development data for optimizing hyperparameters, and then retrain the models on the full training data using the best parameters. For \sysname, we consider domain attributes for 70-year temporal periods and genres, resulting in a total of 21 attributes for each corpus. The numbers of pivot features used in SCL are 4400 and 5048 for the PPCMBE and the PPCEME respectively. The best number of Brown clusters is 200, and the best embedding sizes are 200 and 100 for word2vec and \sysname.

\paragraph{Results} As shown in \autoref{tab:res-ppc}, accuracies are significantly improved by domain adaptation, especially for the PPCEME. English spelling had become mostly uniform and stable since around 1700~\cite{baron2009word}, which may explain why improvements on the PPCMBE are relatively modest, especially in the 1770-1839 epoch. Among the two baseline systems, MEMM performs slightly better than SVM, showing a small benefit to structured prediction. Among the domain adaptation algorithms, \sysname\ clearly outperforms SCL, Brown clustering and word2vec, with an averaged increase of about 0.5\% and 1.5\% accuracies on the PPCMBE and the PPCEME test sets respectively. The metadata attribute information boosts performance by a small but consistent margin, 0.1-0.2\% on average.

\subsection{Adaptation from the Penn Treebank}
Newspaper text is the primary data source for training modern NLP systems. For example, most ``off-the-shelf'' English POS taggers (e.g., the Stanford Tagger~\cite{toutanova2003feature}, SVMTool~\cite{gimenez2004svmtool}, and CRFTagger~\cite{phan2006crftagger}) are trained on the WSJ portion of the Penn Treebank, which is composed of professionally-written news text from 1989. This motivates this evaluation scenario, in which we train the tagger on the Penn Treebank WSJ data and apply it to historical English texts, using all sentences of the PPCMBE and PPCEME for testing.  

\begin{table*}
\centering
\addtolength{\tabcolsep}{-2pt}
\begin{tabular}{l l l l l l l l l}
    \toprule
    \multirow{2}{*}{Target} & \multirow{2}{*}{Normalized} &  \multicolumn{2}{c}{baseline} & \multirow{2}{*}{SCL} & \multirow{2}{*}{Brown} & \multirow{2}{*}{word2vec}\hspace{0.0cm} & \multicolumn{2}{c}{\sysname}\\
    \cmidrule{3-4} \cmidrule{8-9}
    & & SVM & \parbox{1.4cm}{MEMM\\(Stanford)} & & & & \parbox{1.6cm}{single\\embedding} & \parbox{3.4cm}{attribute embeddings\\(error reduction)} \\ \midrule
\textsc{ppcmbe} &  No & 81.12 & 81.35 & 81.66 & 81.65 & 81.75 & 82.34 & \textbf{82.46} (7\%) \\
\textsc{ppceme} &  No & 74.15 & 74.34 & 75.89 & 76.04 & 75.85 & 77.77 & \textbf{77.92} (15\%) \\ 
\textsc{ppceme} & Yes & 76.73 & 76.87 & 77.61 & 77.65 & 77.76 & 78.85 & \textbf{79.05} ($19\%^*$) \\
    \bottomrule
\end{tabular}
\caption{Accuracy results for adapting from the PTB to the PPCMBE and the PPCEME of historical English. $^*$Error reduction for the normalized PPCEME is computed against the unnormalized SVM accuracy, showing total error reduction.}
\label{tab:res-ptb}
\end{table*}

\paragraph{Settings} The feature representations are trained on the union of the PTB and the PPCEME. The domain attributes for \sysname\ are set to include the three corpora themselves (PTB, PPCMBE, and PPCEME), and the genre attributes in the historical corpora. Note that all sentences in the Penn Treebank WSJ data belong to the same genre (news). For SCL, we use the same threshold of $50$ occurrences for pivot features, and include 8089 features that pass this threshold. PTB WSJ sections 19-21 are used for parameter tuning: we find that the best number of Brown clusters is 200, and the optimum embedding sizes are 200 and 100 for word2vec and \sysname.

\paragraph{Spelling normalization} Spelling variants lead to a high percentage of out-of-vocabulary (OOV) tokens in historical texts, which poses problems for POS tagging. We normalize the PPCEME sentences using VARD~\cite{baron2008vard2}, a widely used spelling normalization tool that has been proven to improve performance on POS tagging~\cite{rayson2007tagging} and syntactic parsing~\cite{schneider2014parsing}. VARD is designed specifically for Early Modern English spelling variation, and additional labeled data and training are required for other forms of spelling variation, which we do not consider here.
Following \newcite{schneider2014parsing}, we utilize VARD's auto-normalization function with a 50\% normalization threshold, achieving a balance between precision and recall. At this threshold, a total of 12\% ($236298/1961157$) of the tokens in the PPCEME are normalized.\footnote{We only consider $1:1$ mappings, and ignore 328 normalizations corresponding to $1:n$ mappings.}

\paragraph{Results} As shown in \autoref{tab:res-ptb}, this task is considerably more difficult, with even the best systems achieving accuracies that are nearly 15\% worse than in-domain training.
Nonetheless, domain adaptation can help: \sysname\ improves performance by 1.3\% on the PPCMBE data, and by 3.8\% on the unnormalized PPCEME data. Spelling normalization also helps, improving the baseline systems by more than 2.5\%. The combination of spelling normalization and domain adaptation gives an overall improvement in accuracy from 74.2\% to 79.1\%. The relative error reduction is lower than in the temporal adaptation setting: only 19\% at best, versus 30\% error reduction in temporal adaptation. This is because there are now at least two sources of error --- language change and tagset mismatch --- and unsupervised domain adaptation cannot address mismatches in the tag set.


\section{Analysis}
As expected, the Early Modern English dataset (PPCEME) is considerably more challenging than the Modern British English dataset (PPCMBE): the baseline accuracy is 7\% worse on the PPCEME than the PPCMBE. However, the PPCEME is also more amenable to domain adaptation, with \sysname\ offering considerably larger improvements. One reason is that the PPCEME has many more out-of-vocabulary (OOV) tokens: 23\%, versus 9.2\% in the PPCMBE. Both domain adaptation and normalization help to address this specific issue, and they yield further improvements when used in combination. This section offers further insights on the sources of errors and possibilities for improvement on the PPCEME data.


\subsection{Feature Ablation} \autoref{tab:ablation} presents the results of feature ablation experiments for the non-adapted SVM tagger. Word context features are important for obtaining good accuracies on both IV and OOV tokens. Affix features, particularly suffix features, are crucial for the OOV tokens. The orthographic features are shown to be nearly irrelevant, as long as affix features are present. Overall, the high percentage of OOV tokens can be a major source of errors, as the tagging accuracy on OOV tokens is below 50\% in our best baseline system. Note that these results are for a classification-based tagger; while the Viterbi-based MEMM tagger performs only marginally better overall ($\sim 0.2\%$ improvement), it is possible that its error distribution might be different due to the advantages of structured prediction.

\begin{table} [t]
\centering
\addtolength{\tabcolsep}{-2pt}
\begin{tabular}{l l l l}
    \toprule
    Feature set & IV & OOV & All \\ \midrule
    All features & 81.68 & 48.96 & 74.15  \\[5pt] 
     \, -- word context & 79.69 & 38.62 & 70.23 \\[5pt] 
     \, -- prefix & 81.61 & 46.11 & 73.43 \\
     \, -- suffix & 81.36 & 38.13 & 71.40 \\
     \, -- affix  & 81.22 & 34.40 & 70.44 \\[5pt] 
     \, -- orthographic & 81.68 & 48.92 & 74.14 \\
    \bottomrule
\end{tabular}
\caption{Tagging accuracies of adaptation of our baseline SVM tagger from the PTB to the PPCEME in ablation experiments.}
\label{tab:ablation}
\end{table}

\subsection{Error Analysis} 
The accuracy on out-of-vocabulary (OOV) tokens is generally low, and spelling variation is a major source of OOV tokens. For instance, \example{ye} and \example{thy}, the older forms of \example{the} and \example{your}, are often incorrectly tagged as \postag{NN} and \postag{JJ} in the PPCEME. 
In general, the per-tag accuracies are roughly correlated with the percentages of OOV tokens. Some exceptions including \postag{VB}, \postag{NNP} and \postag{NNS}, where the affix features can be very useful for tagging OOV tokens.

That said, the cross-domain accuracy on in-vocabulary (IV) tokens is also low, at roughly 80\% when adapting from the PTB to the PPCEME. A major source of error here is the mismatch in annotation schemes between the two datasets, which is only partially addressed by a deterministic tag mapping. Table~\ref{tab:tag-acc} presents the SVM accuracy per tag, and the most common error correspondingly. Most of the errors shown in the table are owing to different annotations of the same token in the two corpora. 

One major cause of errors is in misalignments of punctuations and their POS tags. For example, in the PPCEME, 16.6\% of commas are labeled as \postag{.} (sentence-final punctuation), and 12.3\% periods are labeled as \postag{,} (sentence-internal punctuation); these punctuations are less ambiguous in the PTB. The historical corpora lack special tags for colons and ellipses, which are present in the PTB. In contrast to the PTB, there is no distinction between opening quotation mark and closing quotation mark in the PPCEME. \newcite{moon2007part} avoid these difficulties by mapping all the punctuation tokens to a single tag. We did not follow their setting because it would lead to a significant change of test data. However, it should be noted that these ``errors'' are not particularly meaningful for linguistic analysis, and could easily be addressed by heuristic post-processing.

The tagging performance is also impaired by the different annotations of many common words. For example, in the PTB, more than 99.9\% of token \example{to} are labeled as \postag{TO}, but in the PCHE this word can also be labeled as \postag{IN}, distinguishing the infinitive marker from the preposition. The words \example{all}, \example{any} and \example{every} are annotated as quantifiers in the PCHE; this tag is mapped to \postag{JJ}, but these specific words are all labeled as \postag{DT} in the PTB. A simple remapping from \postag{Q} to \postag{DT} leads to an increase of 0.78\% baseline accuracy; it is possible that other changes to the tag mappings of \newcite{moon2007part} might yield further improvements, but a more systematic approach would be outside the bounds of \emph{unsupervised} domain adaptation. 

\begin{table} [t]
\centering
\small
\begin{tabular}{l l l l}
    \toprule
    Tag & \% of OOV & Accuracy & Most common error \\ \midrule
    IN & 6.93 & 82.79 & to/TO \\
    NN & 48.39 & 64.74 & Lord/NNP \\
    DT & 3.45 & 94.62 & that/IN \\
    PRP & 13.57 & 78.80 & other/JJ \\
    , & 0.41 & 87.86 & ./. \\
    JJ & 32.20 & 48.60 & all/DT \\
    CC & 1.98 & 91.29 & for/IN \\
    RB & 26.22 & 65.74 & such/JJ \\
    . & 0.56 & 54.43 & ,/, \\
    VB & 34.69 & 75.06 & have/VBP \\
    NNP & 58.91 & 88.31 & god/NN \\
    NNS & 59.12 & 73.88 & Lords/NNPS \\
    VBD & 25.87 & 81.93 & quoth/NN \\
    VBN & 37.75 & 63.09 & said/VBD \\
    PRP\$ & 13.57 & 85.49 & thy/JJ \\
    \bottomrule
\end{tabular}
\caption{Accuracy (recall) rates per tag with the SVM model, for the 15 most common tags. For each gold category, the most common error word and predicted tag are shown.}
\label{tab:tag-acc}
\end{table}

\subsection{Improvements from Normalization} As shown above, the tagging accuracy decreases from 81.7\% on IV tokens to 49.0\% on OOV tokens. Spelling normalization helps to increase the accuracy by transforming OOV tokens to IV tokens. After normalization, the OOV rate for the PPCEME falls from 23.0\% to 13.5\%, corresponding to a reduction of 41.5\% OOV tokens. Normalization is not perfectly accurate, and the tagging performance for IV tokens drops slightly to 81.2\% on IV tokens. But due to the dramatic decrease in the number of OOV tokens, normalization improves the overall accuracy by more than 2.5\%. We also observe performance drops on tagging OOV tokens after normalization (49.0\% to 48.1\%), which suggests that the remaining unnormalized OOV tokens are the tough cases for both normalization and POS tagging.

\subsection{Improvements from Domain Adaptation} As presented in Table~\ref{tab:adapt-acc}, the tagging accuracies are increased on both IV and OOV tokens with the domain adaptation methods. Compared against the baseline tagger, \sysname-attribute achieves an absolute improvement of 14\% in accuracy on OOV tokens. SCL performs slightly better than Brown clustering and word2vec on IV tokens, but worse on OOV tokens. By incorporating metadata attributes, \sysname-attribute performs  better than \sysname-single on OOV tokens, though the accuracies on IV tokens are similar. Interestingly, the venerable method of Brown clustering (slightly) outperforms both word2vec and SCL.

\begin{table}
\centering
\addtolength{\tabcolsep}{-2pt}
\begin{tabular}{l l l l}
    \toprule
    System & IV & OOV & All \\ \midrule
    SVM & 81.68 & 48.96 & 74.15  \\[4pt]
    SCL & 82.01 & 55.45 & 75.89 \\
    Brown & 81.81 & 56.76 & 76.04 \\
    word2vec & 81.79 & 56.00 & 75.85 \\
    \sysname-single  & 82.30 & 62.63 & 77.77 \\
    \sysname-attribute & 82.34 & 63.16 & 77.92 \\
    \bottomrule
\end{tabular}
\caption{Tagging accuracies of domain adaptation models from the PTB to the PPCEME.}
\label{tab:adapt-acc}
\end{table}

We further study the relationship between domain adaptation and spelling normalization by looking into the errors corrected by both approaches. Domain adaptation yields larger improvements than spelling normalization on both IV and OOV tokens, although as noted above, the approaches are somewhat complementary. The results show that among the 60,928 error tokens corrected by VARD, 60\% are also corrected by \sysname-attribute, while the remaining 40\% would be left uncorrected by the domain adaptation technique. Conversely, among the errors corrected by \sysname-attribute, 38\% are also corrected by VARD, while the remaining 62\% would be left uncorrected. The overlap of reduced errors is because both approaches exploit similar sources of information, including affixes and local word contexts. 



\section{Related Work}

\paragraph{Domain adaptation} Early work on domain adaptation focuses on supervised setting, in which some amount of labeled instances are available in the target domain~\cite{jiang2007instance,daume2007frustratingly,finkel2009hierarchical}. Unsupervised domain adaptation is more challenging but attractive in many applications, and several representation learning methods have been proposed for addressing this problem. Structural Correspondence Learning~\cite[SCL]{blitzer2006domain} and marginalized denoising autoencoders~\cite[mDA]{chen2012marginalized} seek cross-domain representations that are useful to predict a subset of features in the original instances, called pivot features. \newcite{schnabel2014flors} directly induce distributional representations for POS tagging based on local left and right neighbors of the token.
 More recent work trains cross-domain representations with neural networks, with additional objectives such as minimizing errors in the source domain and maximizing domain confusion loss~\cite{ganin2015unsupervised,tzeng2015simultaneous}. We show the Feature Embedding model, which is specifically designed for NLP problems with feature templates~\cite{yang2015unsupervised}, achieves strong performance on historical adaptation tasks.

\paragraph{Historical texts} Historical texts differ from modern texts in spellings, syntax and semantics, posing significant challenges for standard NLP systems, which are usually trained with modern news text. Numerous resources have been created for overcoming the difficulties, including syntactically annotated corpora~\cite{kroch2004penn,kroch2010penn,galves2010tycho} and spelling normalization tools~\cite{giusti2007automatic,baron2008vard2}. Most previous work focuses on normalization, which can significantly increase tagging accuracy on historical English~\cite{rayson2007tagging} and German~\cite{scheible2011evaluating}. Similar improvements have been obtained for syntactic parsing~\cite{schneider2014parsing}. Domain adaptation offers an alternative approach which is more generic --- for example, it can be applied to any corpus without requiring the design of a set of normalization rules. As shown above, when normalization is possible, it can be combined with domain adaptation to yield better performance than that obtained by either approach alone.

\section{Conclusion}
Syntactic analysis is a key first step towards processing historical texts, but it is confounded by changes in spelling and usage over time. We empirically evaluate several unsupervised domain adaptation approaches for POS tagging of historical English texts. We find that domain adaptation methods significantly improve the tagger performance on two historical English treebanks, with relative error reductions of 30\% in the temporal adaptation setting. \sysname\ outperforms other domain adaptation approaches, showing the importance of adapting the entire feature vector, rather than simply using word embeddings. Normalization and domain adaptation combine to yield even better performance, with a total of 5\% raw accuracy improvement over a baseline classifier in the most difficult setting. 
Error analysis reveals that tagset mismatch is the most common source of errors for in-vocabulary words. We hope that our work encourages further research on domain adaptation for historical texts and provides useful baselines in these efforts.

\paragraph{Acknowledgments}
This research was supported by National Science Foundation award 1349837, and by the National Institutes of Health under award number R01GM112697-01. We thank the reviewers for their helpful feedback.

\bibliographystyle{naaclhlt2016}
\bibliography{cite-strings,cites,cite-definitions}

\pagebreak

\onecolumn
\begin{appendices}
\section{Appendix: Tag Mappings}
\label{app:mappings}

\vspace{.2cm}

This table provides the full mapping from Penn-Corpus of Historical English tags to Penn Treebank Tags used in our evaluation.

\vspace{12pt}

\noindent
\addtolength{\tabcolsep}{-2pt}
\begin{tabular}{p{.4\textwidth}p{.3\textwidth}p{.3\textwidth}}
    PCHE $\rightarrow$ PTB & PCHE $\rightarrow$ PTB & PCHE $\rightarrow$ PTB \\ \hline
    \postag{,} (sent-internal) $\rightarrow$ \postag{,} (comma) & \postag{ELSE} $\rightarrow$ \postag{RB} & \postag{OTHER} $\rightarrow$ \postag{PRP}\\
    \postag{.} (sent-final) $\rightarrow$ \postag{.} (sent-final) & \postag{EX} $\rightarrow$ \postag{EX} & \postag{OTHER\$} $\rightarrow$ \postag{PRP}\\
    \postag{'} (single quote) $\rightarrow$ \postag{''} (closing quote) & \postag{FOR} $\rightarrow$ \postag{IN} & \postag{OTHERS\$} $\rightarrow$ \postag{PRP}\\
    \postag{\"} (double quote) $\rightarrow$ \postag{''} (closing quote) & \postag{FP} $\rightarrow$ \postag{CC} & \postag{OTHERS} $\rightarrow$ \postag{PRP}\\
    \postag{\$} $\rightarrow$ \postag{PRP\$} & \postag{FW} $\rightarrow$ \postag{FW} & \postag{P} $\rightarrow$ \postag{IN}\\
    \postag{ADJ} $\rightarrow$ \postag{JJ} & \postag{HAG} $\rightarrow$ \postag{VBG} & \postag{PRO} $\rightarrow$ \postag{PRP}\\
    \postag{ADJR} $\rightarrow$ \postag{JJR} & \postag{HAN} $\rightarrow$ \postag{VBN} & \postag{PRO\$} $\rightarrow$ \postag{PRP\$}\\
    \postag{ADJS} $\rightarrow$ \postag{JJS} & \postag{HV} $\rightarrow$ \postag{VB} & \postag{Q} $\rightarrow$ \postag{JJ}\\
    \postag{ADV} $\rightarrow$ \postag{RB} & \postag{HVD} $\rightarrow$ \postag{VBD} & \postag{QS} $\rightarrow$ \postag{RBS}\\
    \postag{ADVR} $\rightarrow$ \postag{RBR} & \postag{HVI} $\rightarrow$ \postag{VB} & \postag{QR} $\rightarrow$ \postag{RBR}\\
    \postag{ADVS} $\rightarrow$ \postag{RBS} & \postag{HVN} $\rightarrow$ \postag{VBN} & \postag{RP} $\rightarrow$ \postag{RB}\\
    \postag{ALSO} $\rightarrow$ \postag{RB} & \postag{HVP} $\rightarrow$ \postag{VBP} & \postag{SUCH} $\rightarrow$ \postag{RB}\\
    \postag{BAG} $\rightarrow$ \postag{VBG} & \postag{INTJ} $\rightarrow$ \postag{UH} & \postag{TO} $\rightarrow$ \postag{TO}\\
    \postag{BE} $\rightarrow$ \postag{VB} & \postag{MD} $\rightarrow$ \postag{MD} & \postag{VAG} $\rightarrow$ \postag{VBG}\\
    \postag{BED} $\rightarrow$ \postag{VBD} & \postag{N} $\rightarrow$ \postag{NN} & \postag{VAN} $\rightarrow$ \postag{VBN}\\
    \postag{BEI} $\rightarrow$ \postag{VB} & \postag{N\$} $\rightarrow$ \postag{NN} & \postag{VB} $\rightarrow$ \postag{VB}\\
    \postag{BEN} $\rightarrow$ \postag{VBN} & \postag{NEG} $\rightarrow$ \postag{RB} & \postag{VBD} $\rightarrow$ \postag{VBD}\\
    \postag{BEP} $\rightarrow$ \postag{VBZ} & \postag{NPR} $\rightarrow$ \postag{NNP} & \postag{VBI} $\rightarrow$ \postag{VB}\\
    \postag{C} $\rightarrow$ \postag{IN} & \postag{NPR\$} $\rightarrow$ \postag{NNP} & \postag{VBN} $\rightarrow$ \postag{VBN}\\
    \postag{CONJ} $\rightarrow$ \postag{CC} & \postag{NPRS} $\rightarrow$ \postag{NNPS} & \postag{VBP} $\rightarrow$ \postag{VBP}\\
    \postag{D} $\rightarrow$ \postag{DT} & \postag{NPRS\$} $\rightarrow$ \postag{NNPS} & \postag{WADV} $\rightarrow$ \postag{WRB}\\
    \postag{DAG} $\rightarrow$ \postag{VBG} & \postag{NS} $\rightarrow$ \postag{NNS} & \postag{WARD} $\rightarrow$ \postag{VB}\\
    \postag{DAN} $\rightarrow$ \postag{VBN} & \postag{NS\$} $\rightarrow$ \postag{NNS} & \postag{WD} $\rightarrow$ \postag{WDT}\\
    \postag{DO} $\rightarrow$ \postag{VB} & \postag{NUM} $\rightarrow$ \postag{CD} & \postag{WPRO} $\rightarrow$ \postag{WP}\\
    \postag{DOD} $\rightarrow$ \postag{VBD} & \postag{NUM\$} $\rightarrow$ \postag{CD} & \postag{WPRO\$} $\rightarrow$ \postag{WP\$}\\
    \postag{DOI} $\rightarrow$ \postag{VB} & \postag{ONE} $\rightarrow$ \postag{PRP} & \postag{WQ} $\rightarrow$ \postag{IN}\\
    \postag{DON} $\rightarrow$ \postag{VBN} & \postag{ONES} $\rightarrow$ \postag{PRP} & \postag{X} $\rightarrow$ \postag{X}\\
    \postag{DOP} $\rightarrow$ \postag{VBP} & \postag{ONE\$} $\rightarrow$ \postag{PRP\$}\\
\end{tabular}
\end{appendices}

\end{document}